\documentclass[letterpaper]{article} 
\usepackage{aaai23}  
\usepackage{times}  
\usepackage{helvet}  
\usepackage{courier}  
\usepackage[hyphens]{url}  
\usepackage{graphicx} 
\usepackage{natbib}  
\usepackage{caption} 
\usepackage{algorithm}
\usepackage{algorithmic}

\usepackage{booktabs}
\usepackage{multirow}
\usepackage{makecell}

\usepackage{newfloat}
\usepackage{listings}
\DeclareCaptionStyle{ruled}{labelfont=normalfont,labelsep=colon,strut=off} 
\floatstyle{ruled}
\newfloat{listing}{tb}{lst}{}
\floatname{listing}{Listing}
\nocopyright
\title{See Your Heart: Psychological states Interpretation through Visual Creations}
\author {
    Likun Yang,\textsuperscript{\rm 1,2}
    Xiaokun Feng, \textsuperscript{\rm 1,2}
    Xiaotang Chen, \textsuperscript{\rm 1}
    Shiyu Zhang, \textsuperscript{\rm 1}
    Kaiqi Huang \textsuperscript{\rm 1}
}
\affiliations {
    \textsuperscript{\rm 1} Institute of Automation, Chinese Academy of Sciences\\
    \textsuperscript{\rm 2} University of Chinese Academy of Sciences, School of Artificial Intelligence\\
}

\begin{document}

\maketitle

    \begin{abstract}
        In psychoanalysis, generating interpretations to one's psychological state through visual creations is facing significant demands. The two main tasks of existing studies in the field of computer vision, sentiment/emotion classification and affective captioning, can hardly satisfy the requirement of psychological interpreting. To meet the demands for psychoanalysis, we introduce a challenging task, \textbf{V}isual \textbf{E}motion \textbf{I}nterpretation \textbf{T}ask (VEIT). VEIT requires AI to generate reasonable interpretations of creator's psychological state through visual creations. To support the task, we present a multimodal dataset termed SpyIn (\textbf{S}and\textbf{p}la\textbf{y} \textbf{In}terpretation Dataset), which is psychological theory supported and professional annotated. Dataset analysis illustrates that SpyIn is not only able to support VEIT, but also more challenging compared with other captioning datasets. Building on SpyIn, we conduct experiments of several image captioning method, and propose a visual-semantic combined model which obtains a SOTA result on SpyIn. The results indicate that VEIT is a more challenging task requiring scene graph information and psychological knowledge. Our work also show a promise for AI to analyze and explain inner world of humanity through visual creations.
    \end{abstract}

    \section{1. Introduction}

    \begin{figure}[t]
        \centering
        \includegraphics[width=0.9\columnwidth]{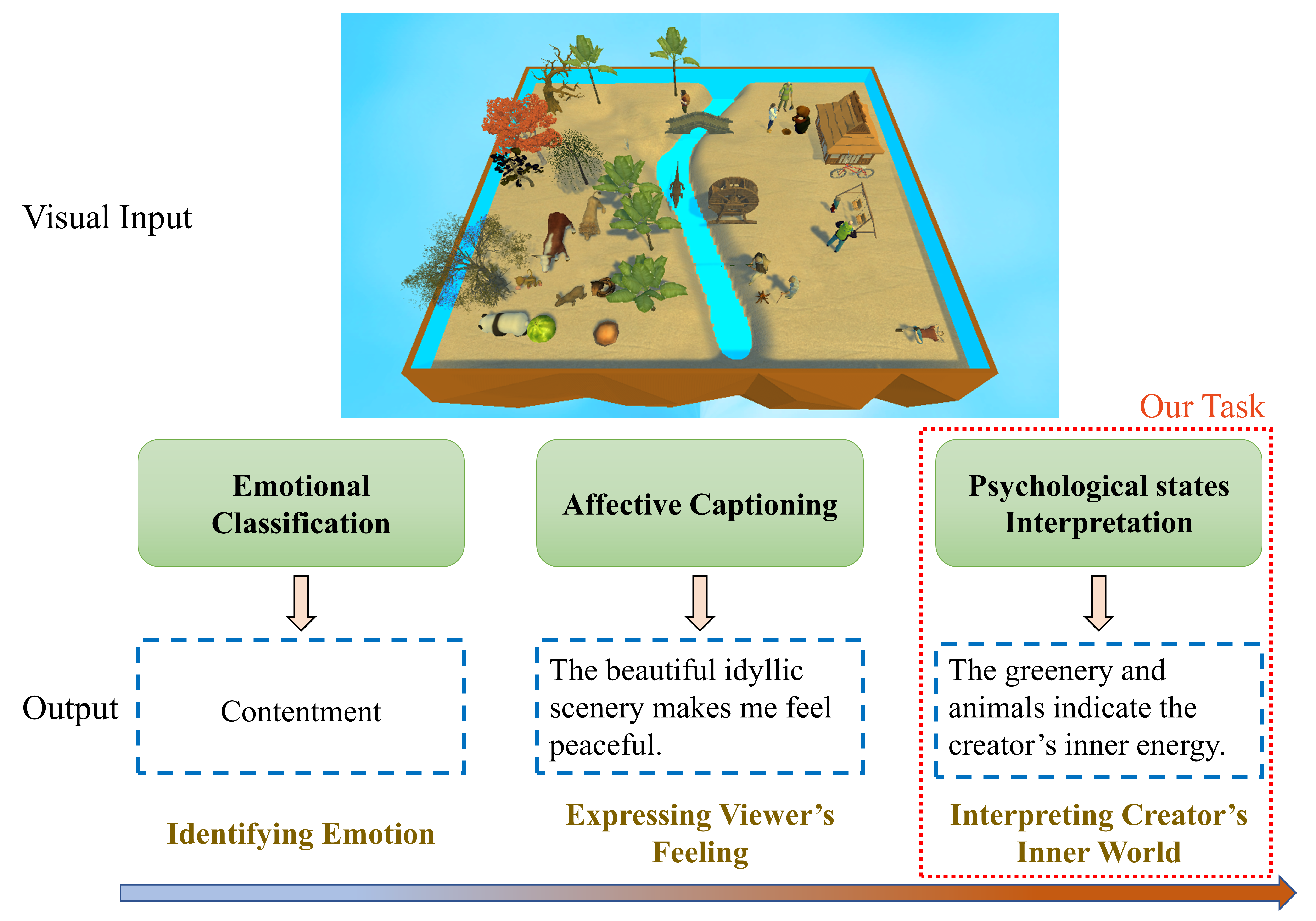} 
        \caption{\textbf{Difference between emotion classification, affective captioning and psychological states interpretation.} We address the Visual Emotion Interpretation Task (VEIT) is to generate reasonable interpretations of the creator's psychological state embedded in visual creations.}
        \label{fig1}
    \end{figure}

    \begin{figure*}[t]
        \centering
        \includegraphics[width=0.9\textwidth]{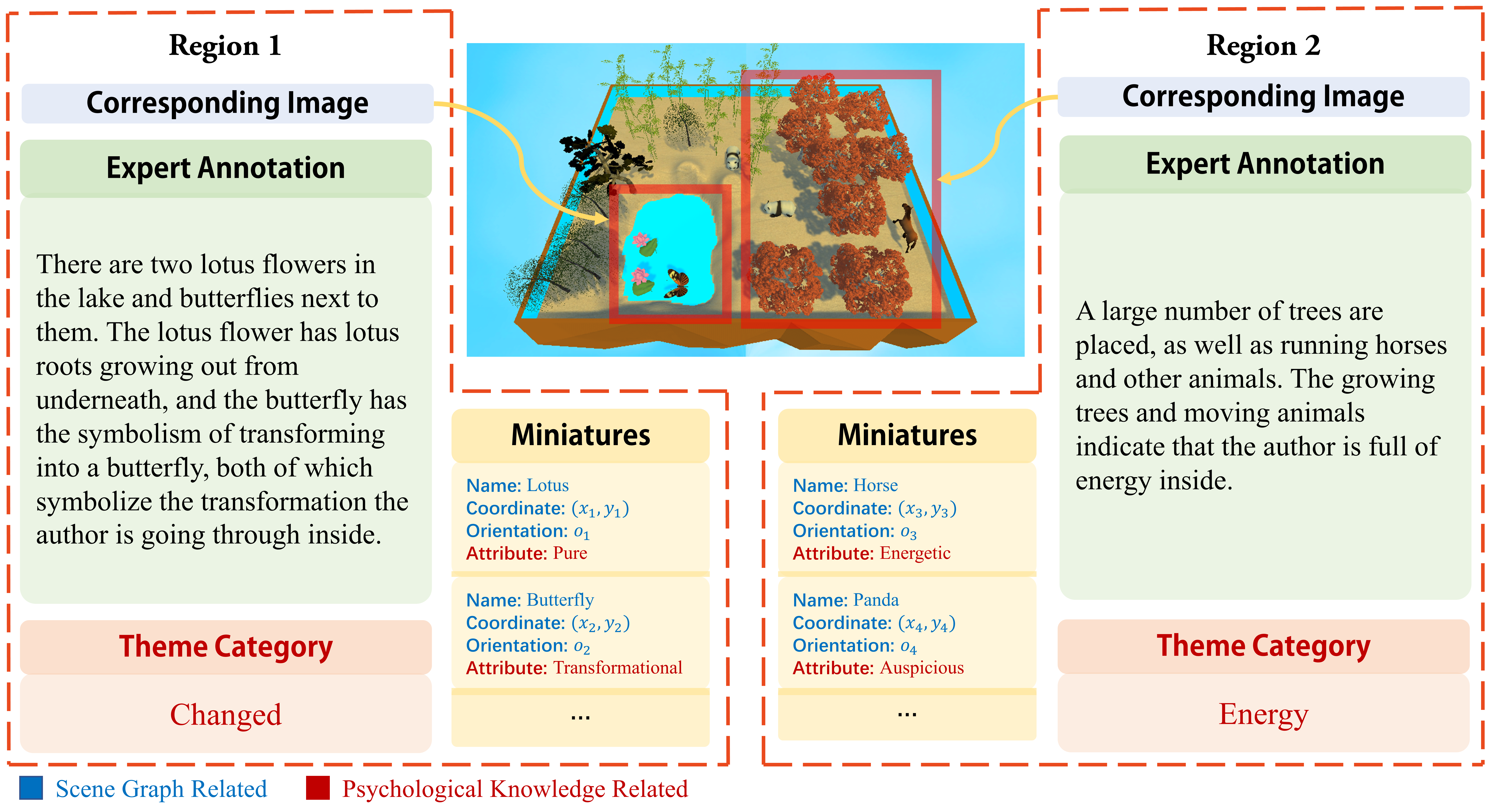} 
        \caption{\textbf{Example regions of SpyIn.} Experts frame expressive regions, interpret the creator's psychological state through the specific region, categorize them with themes. SpyIn provides two kinds of extra information, including scene graph related information and psychological knowledge related information.}
        \label{fig2}
    \end{figure*}
    
    One of the most common ways for human nature to express oneself is through creating works (like artworks, articles, music, etc.). Visual creating has been proven as an important way to reflect one's psychological state \cite{way_to_reflect_inner_world_1,way_to_reflect_inner_world_2}. Thus, it is important to interpret one's psychological state through visual creations, which is also commonly used in the fields of artworks appreciation \cite{Inner_interpret_on_art_appreciated_1,Inner_interpret_on_art_appreciated_2}, clinical psychoanalysis \cite{visual_interpretion_on_clinical_pys_1,visual_interpretion_on_clinical_pys_2}, etc. Particular in psychoanalysis, interpreting one's psychological state through visual creations is an important psychological assessment method facing significant demands \cite{img_iner_use_1,img_iner_use_2}.
    
    To the best of our knowledge, there are two main types of works in computer vision field that combine vision with psychology, which are sentiment/emotion classification \cite{emoclass_1,emoclass_2,emoclass_3} and affective captioning \cite{senticap,artemis,itsokay}. Sentiment/emotion classification treats sentiment/emotion analysis as a classification task, which stay at the ability of identifying sentiment/emotion but not generating sentences to interpret. Affective captioning focuses on generating captions of images with specific emotional tendency. Notably, ArtEmis \cite{artemis}, as a cutting-edge work in affective captioning, annotates descriptions of the viewers' feeling based on visual artworks. Affective captioning is capable of expressing emotion, however, the emotion they expressing is from themselves rather than emotion of others (the visual work creators). In short, affective captioning task “see” the heart of viewers but not the heart of creators (See Figure 1). As the increasing demands for psychoanalysis \cite{visual_interpretion_on_clinical_pys_1,img_iner_use_2}, AI that is able to interpret creator's psychological state through their visual creations, is in need. However, the existing tasks in computer vison field can hardly meet the demands. Therefore, we propose a new task, \textbf{V}isual \textbf{E}motion \textbf{I}nterpretation \textbf{T}ask (VEIT). VEIT is to generate reasonable interpretations of the creator's psychological state embedded in visual creations. Notably, the word “emotion” in VEIT does not refer to specific emotions, but to the psychological state \cite{psystat} of creators.
    
    In this paper, we present a psychological theory supported and professional annotated dataset, SpyIn (\textbf{S}and\textbf{p}la\textbf{y} \textbf{In}terpretation Dataset), to support VEIT. SpyIn is a challenging dataset and tightly fit the requirement of VEIT in following aspects.

    \subsubsection{Visual sandplay works.} A Sandplay work is a scene created by the client to show a picture of his/her inner world with different varieties of miniatures \cite{visual_interpretion_on_clinical_pys_2}. We focus on sandplay works for two reasons. Firstly, Sandplay is an important psychoanalytical tool supported by psychological theory \cite{visual_interpretion_on_clinical_pys_2,img_iner_use_1} ensuring the reasonable connection between the visual work and the psychological state of creators, which satisfies the requirement of VEIT. Secondly, sandplay works are widely used in psychotherapy \cite{img_iner_use_2}. The realistic applications not only enable us to access broad samples and professional annotation, but also give a prospective application in psychological diagnosis supporting.

    \subsubsection{Expert annotations.} As the first dataset serving VEIT, SpyIn has professional annotating team and rigorous annotating process, ensuring the interpretation of the visual work being reasonable and accurate. We engage expert psychotherapists to annotate the collected sandplay works. The annotating processes are not only in line with psychotherapeutic standards but also supported by psychometric scales (See Section 3).  Furthermore, we adopt annotations from expert psychotherapists rather than from creators themselves for following two reasons. Firstly, the true feeling or psychological state may not always be willing to be expressed or accurately articulated by creators themselves. Secondly, we need to make sure that the annotators are on the same page with the AI served by the dataset, which they are both interpreting the psychological states of the creators from the third-person perspective.

    \subsubsection{Challenging dataset severing for challenging task.} According to dataset analysis (See Section 3), SpyIn is not only psychological canonical but also richer and more abstract on language, more diverse on visual content and emotion. Comparing to existing captioning datasets, SpyIn dataset is more challenging, which also indicats the difficulty of VEIT. Both the statistic of dataset and experiment results show that VEIT relies on further knowledge rather than only on visual input, which also confirms the challenging of VEIT.

    \subsubsection{Multi-Modality information providing.} Because of the challenging nature of VEIT, classical image captioning models may not perform well (See Section 5). To offer possibilities for a wide variety of models, we provide not only the visual images and the corresponding interpretation annotation but also multimodal information. The multimodal information including scene graph related information, such as the location coordinates of the miniatures, orientation of the miniatures, height map of the scene and etc. The multimodal information also including psychological knowledge information, such as the theme categories (i.e. a standard categorization method in psychological theory \cite{themes_1,themes_2}) and attribute of the miniatures. The multimodal information we provided offers possibilities for variety of approaches to deploy on SpyIn, such as scene graph generation \cite{scene_graph}, semantic information embedding \cite{semantic_embedding}, etc.
    
    We conduct experiments on SpyIn for VEIT. We not only benchmark several representative image captioning approaches, but also perform a visual-semantic combined method achieved SOTA in SpyIn. Compared with existing methods in image captioning, our approach makes good use of the multimodal information rather than visual features only, which meets the further knowledge requirement of VEIT. 
    
    The contributions of our work are summarized as follows:
    
    \begin{itemize}
        \item To meet the increasing demand of psychoanalysis, we define a challenging task, \textbf{V}isual \textbf{E}motion \textbf{I}nterpretation \textbf{T}ask (VEIT), requiring AI to generate reasonable interpretation of the creators' psychological states embedded in visual creations.
        \item To support VEIT, we present a challenging multimodal dataset, SpyIn, which is psychological theory supported and professional annotated.
        \item Building on VEIT and SpyIn, we benchmark the performance of several image captioning approaches. We also provide a visual-semantic combined model leveraging multimodal information and achieve SOTA on SpyIn. The results show the promise for AI to analyze and articulate the psychological states of humanity through our visual works.
    \end{itemize}

    \section{2. Related work}
    
    \subsection{2.1. Vision combines emotion}
    Vision combined emotion is a recently popular research area. The early works related is emotional semantic image retrieval \cite{retrieval_emo_1,retrieval_emo_2}, which make connections between low-level image features and emotions with the aim to perform automatic image retrieval and categorisation. As the progresses in this area, there are two main tasks: sentiment/emotional classification \cite{emoclass_2,emoclass_3} and affective captioning\cite{affective_cap,senticap,artemis,itsokay}. Emotional classification treats emotions as an image classification problem, which remaining on identifying emotions rather than express them with narrative sentences. Affective captioning expresses the feelings of the viewers, but can hardly analyze and interpret the psychological state of the creators.
    
    \subsubsection{Emotional classification.}
    Most existing works in Computer Vision treat emotions into classification task. Their visual inputs are most based on real world images \cite{photo_base_1,emoclass_3}, and a small number of works are based on visual creations such as paintings \cite{art_emo_class}. Visual classification uses different categories to represent emotions, such as two or three polarity levels \cite{pollar_class}, different level emotional scheme \cite{emotionclass_scale}, etc. Although there are many efforts to classify emotions, they are difficult to express them in statements.
    
    \subsubsection{Affective captioning.}
    There is particular rare work both involved emotion and captioning. Mathews \cite{senticap} and follow-ups like \cite{affective_cap} are early works focused on injecting two emotions (positive and negative) into image captioning. They are works applied on photograph-based images which can hardly reflect the creators' emotion. The only two emotions injecting to the captioning is based on the visual cue through objective things without analyze of the scene. The recent work ArtEmis \cite{artemis} and its follow work \cite{itsokay} proposed dataset combined emotional information and visual artworks. But the emotion information is the viewer's feeling, but not focused on interpreting the creators' psychological state. Comparing to previous work, VEIT concentrating on interpret the creator's psychological state.
    
    \subsection{2.2. Image captioning}
    VEIT outputs the interpretation of creators' psychological state, which is similar to image captioning \cite{show_tell} task in terms of the output form. Therefore, we build our baselines on image captioning techniques.
    
    \subsubsection{Image captioning method.}
    As the development of deep neural network, there are a lot of deep-net approaches for image captioning\cite{self_crit,bottomup,self_n,attonatt}. We apply several “standard” approaches on our dataset. Vinyals etc. \cite{show_tell} introduced CNN plus LSTM network into image captioning. Xu etc. \cite{SAT} introduced LSTM with attention into image caption. Get rid of the CNN-LSTM structure, Cornia etc. \cite{M2} gains a good result applying transformer structure into image captioning.
    The work of Xu etc. \cite{SAT} and Cornia etc. \cite{M2} are SOTA image captioning approaches of CNN-LSTM architecture and transformer architecture respectively. We choose these two representative methods as baselines on SpyIn.

    \subsubsection{Image captioning dataset.}
    For a long time, researchers have devoted to the task of image captioning as a bridge from vision to language, which has led a great deal of work on datasets in particular \cite{flickr8k,flickr30k, generation_1,conceptual,coco,artemis}. Among them, COCO \cite{coco}, as the most commonly used dataset on image captioning, is a large-scale dataset with over 330K natural images. Notably, ArtEmis \cite{artemis}, as the cutting-edge work on affective captioning dataset, connects emotions to visual arts captioning. The annotations of ArtEmis are descriptions to viewer's emotions on visual stimuli of artworks. There are two main differences between SpyIn and ArtEmis. Firstly, ArtEmis subjectively express the feeling of viewers, but SpyIn objectively assess the psychological state of creators. Secondly, ArtEmis is annotated by non-professional viewers, but SpyIn is annotated with psychological expertise. Comparing with ArtEmis, SpyIn is a more professional dataset serving for a more professional task, to meet the demands of psychoanalysis.
    
    \section{3. Sandplay interpretation dataset}
    In total, SpyIn containing 4K high-quality pairs of sandplay region images and corresponding expert annotations. For each sandplay image, we also provide corresponding multimodal information for optional input (Section 3.1). SpyIn is constructed to serve the VEIT, requiring both the visual images and the annotations reasonably connect with creators' psychological states. To meet the requirements, we not only conduct a more psychological rigorous and professional data collection process than existing captioning datasets, but also perform a psychological accuracy analysis to validate (Section 3.1). Furthermore, SpyIn is a richer and more challenging dataset than existing image captioning datasets. SpyIn is richer and more abstract in language, more diverse in image content, and more complex in categories (Section 3.2 and 3.3).

    \subsection{3.1. Data collection and validation}
    \subsubsection{Visual image collection.}
    We invited 5000 clients to join the process of creating the sandplay works. Each client was also asked to finished an SCL-90 psychological scale (i.e. Symptom Checklist-90 \cite{scl90}) after the completion of their creation. All the subjects involved this process were paid for their labor. Among the 5000 original sanplay works, we manually eliminated the casually created sandplay works and reserved high-quality created sandplay works for annotation.
    
    \subsubsection{Experts annotation.}
    To ensure the annotation to each sandplay work can psychological accurately interpret the creators' psychological state, we engaged 5 expert psychotherapists to annotate the collected sandplay works. It took at least 40 minutes for the annotating process of each sandplay work. To demonstrate the high quality of our annotation, in Section 3.2, we compared the annotation of each sandplay work with the corresponding SCL-90 psychological scale, which validated the psychological accuracy of our annotation.
    
    \subsubsection{Multimodal information providing.}
    For each sandplay works, the expert psychotherapists first framed the expressive regions, then analyzed the specific psychological state expressed with these regions. For each region, the experts would classify it into a theme category, which is a standard categorization method in psychological theory \cite{themes_1,themes_2}. After that, the experts would also annotate each region with the interpretation of the creator's psychological state. For each region, all miniatures appeared are automatically record by our annotation tool and output the name, location coordinates, orientation, psychological attribute of the them. A height map to reflecting mountains and lakes is also provided by our annotation tool.
    
    Besides the interpretations of experts, we grouped this extra information into two types: scene graph related information and psychological knowledge information. Scene graph related information including the name, location coordinates, orientation of miniatures and height map of the scene. Psychological knowledge information including the theme categories, psychological attribute of miniatures.
    
    \subsubsection{Psychological accuracy validation.}
    According to visual creating collection process, after each client creats the sandplay works, they are also asked to fill out an SCL-90 psychological scale. The Symptom Checklist-90 (SCL-90) is one of the world's most well-known mental health test scales and is widely used in evaluating psychological states\cite{scl90}. We adopt the conventional division of SCL-90, using a total score of 160 as the cut-off value, where regards clients below the value as normal, otherwise as abnormal. As for the sandplay work annotation, we regard the sandplay work which was annotated by experts with wounding themes (i.e. a group of negative themes) as abnormal and rest as normal. A result comparing of the SCL-90 and expert annotation is shown in table 1. According to psychological standard \cite{psy_accuracy}, with an accuracy rate of more than 70\% and false alarm rate of less than 25\%, the annotation of the sandplay works is in line with psychological assessment standards. The results validate the psychological consistency between expert interpretations and the creators' psychological state.
    
    \begin{table}[thb]
        \centering
        \resizebox{.95\columnwidth}{!}{
        \begin{tabular}{cccccc}
            \toprule
            Total accuracy & Positive accuracy & Negtive accuracy & FPR &  FNR  \\
            \midrule
            \textbf{0.77} & 0.54 & 0.84 & \textbf{0.16} & 0.01 \\
            \bottomrule
        \end{tabular}
        }
        \caption{\textbf{Annotation result compare with SCL-90.} The accuracy include the positive accuracy(abnormal subjects accuracy), and negative accuracy(normal subjects accuracy). FPR means false positive rate, FNR means false negative rate.}
        \label{table1}
    \end{table}

    \subsection{3.2. Linguistic analysis}

    \begin{table}[thb]
        \centering
        \resizebox{.95\columnwidth}{!}{
        \begin{tabular}{lccccccc}
            \toprule
            Dataset & Words & Sentence & Nouns &  Verbs & Adjectives & $c_{score}$ & $s_{score}$ \\
            \midrule
            SpyIn & \textbf{17.4} & \textbf{2.3} & \textbf{8.6} & \textbf{5.1} & \textbf{3.2} & \textbf{2.574} & \textbf{0.988} \\
            ArtEmis & 15.9 & 1.3 & 4.0 & 3.0 & 1.6 & 2.622 & 0.982\\
            COCO Captions & 10.5 & 1.1 & 3.7 & 1.2 & 0.8 & 3.163 & 0.920\\
            Conceptual Capt. & 9.6 & 1.2 & 3.8 & 1.1 & 0.9 & 3.026 & 0.954 \\
            Flickr30k Ent. & 12.3 & 1.2 & 4.2 & 1.8 & 1.1 & 3.180 & 0.904 \\
            Google Refexp & 8.4 & 1.0 & 3.0 & 0.8 & 1.0 & 3.208 & 0.902 \\
            \bottomrule
        \end{tabular}
        }
        \caption{\textbf{Linguistic indicators of SpyIn vs. previous works.} Average num of words, sentences, nouns, verbs and adjectives indicates SpyIn is most richness in language, $c_{score}$ indicates SpyIn has the most abstractness, $s_{score}$ indicates SpyIn has highest ratio of affective sentences.}
        \label{table2}
    \end{table}

    \subsubsection{Richness analysis.}
    We measure the linguistic richness through syntax analysis in terms of the average num of words, sentences, nouns, verbs and adjectives. Compared with existing captioning datasets, as shown in Table 1, SpyIn achieves the maximum value in all the above indicators, which indicates the richness of the expert annotation of SpyIn.
    
    \subsubsection{Abstractness analysis.}
    Interpretation of pychological states often relies more on abstract description. For measuring the abstractness or concreteness, we use the lexicon in Brysbaert et al. \cite{atractness} which provides for 40,000 word lemmas a rating from 1 to 5 reflecting their concreteness. For instance, apple and ball are maximally concrete/tangible objects, getting a score of 5, but happy and idealistic are quite abstract (with scores 2.56 and 1.21, resp.). We take the average of scores corresponding to each word in a sentence as the concreteness score of the sentence. By averaging the concreteness score of each sentence in the dataset, we get the value (denoted as $c_{score}$ ) reflecting the concreteness degree of this dataset. The $c_{score}$ of SpyIn is lower than the other existing captioning datasets (see Table 2). In other words, SpyIn contains more abstract references.
    
    \subsubsection{Sentiment analysis.}
    In addition to being rich and abstract, SpyIn also contains more sentences with sentiment. We use a RoBERTa \cite{senti_ratio} based sentiment analyzer model (trained with SemEval 2017 corpus \cite{roberta_pretrain}) to demonstrate this. The model output $y$, and we judge specific sentence is affective if $abs(y)>0.7$. We analyze each annotation in the dataset, and judge whether it is a affective sentence. Thus, we can calculate the ratio of affective sentences (denoted as $s_{ratio}$) in each dataset (see Table 2). The results show that the SpyIn dataset has the highest $s_{ratio}$.
    
    \subsection{3.3. Multimodal analysis}
    
    \subsubsection{Visual content analysis.}
    Comparing to most common used and photograph based captioning dataset COCO \cite{coco}, a sandplay work usually containing a larger scene with more visual entities. Each sandplay contains an average of 31.9 visual entities, while COCO contains 7.36 visual entities per image. In addition, sandplay works containing more diverse variety of visual entities. There are 494 categories of visual entities containing in SpyIn, far higher than the 80 categories in COCO. The richness and diverse on visual entities illustrates the more complex scenes in sandplay works, which brings more visual information but also more challenges to SpyIn.
    
    \subsubsection{Pychological state categories analysis.}
    Follow the standard of psychological theory \cite{themes_1,themes_2}, we use themes to categorize psychological states. Our dataset categorized the regions into 24 themes, including two main categories: Healing themes and Wounding themes. On average, each sandplay work contains 8.09 categories of 24 themes, while the Artemis sample contains 2.91 categories of 8 emotional labels. This indicates that SpyIn not only has more categorize to distinguish different samples, but also has richer psychological information for each visual work.
    
    The distribution of theme categories is balanced with visual works. After a fair normalization, the variance of the categories distribution with visual works of SpyIn is 0.402, while the ArtEmis is 0.603. 
    
    Notably, the psychological states relies rarely on visual entities, which means the same visual entity may reflect different psychological state in different scenes, instead of being fixed. In order to visually show the relationship between psychological state and visual entities, we encode theme categories and visual entity categories into vectors and perform a PCA. See Figure 3, the spatial distribution of each theme is mixed, which indicates that different kinds of psychological states distributed evenly with visual entity categories. Specifically, it indicates that the psychological states is not strongly relies on the appearance of specific visual entities. This phenomenon illustrates that the psychological states relies weakly on local visual features, but may relies on further information.
    
    \begin{figure}[htb]
        \centering
        \includegraphics[width=0.8\columnwidth]{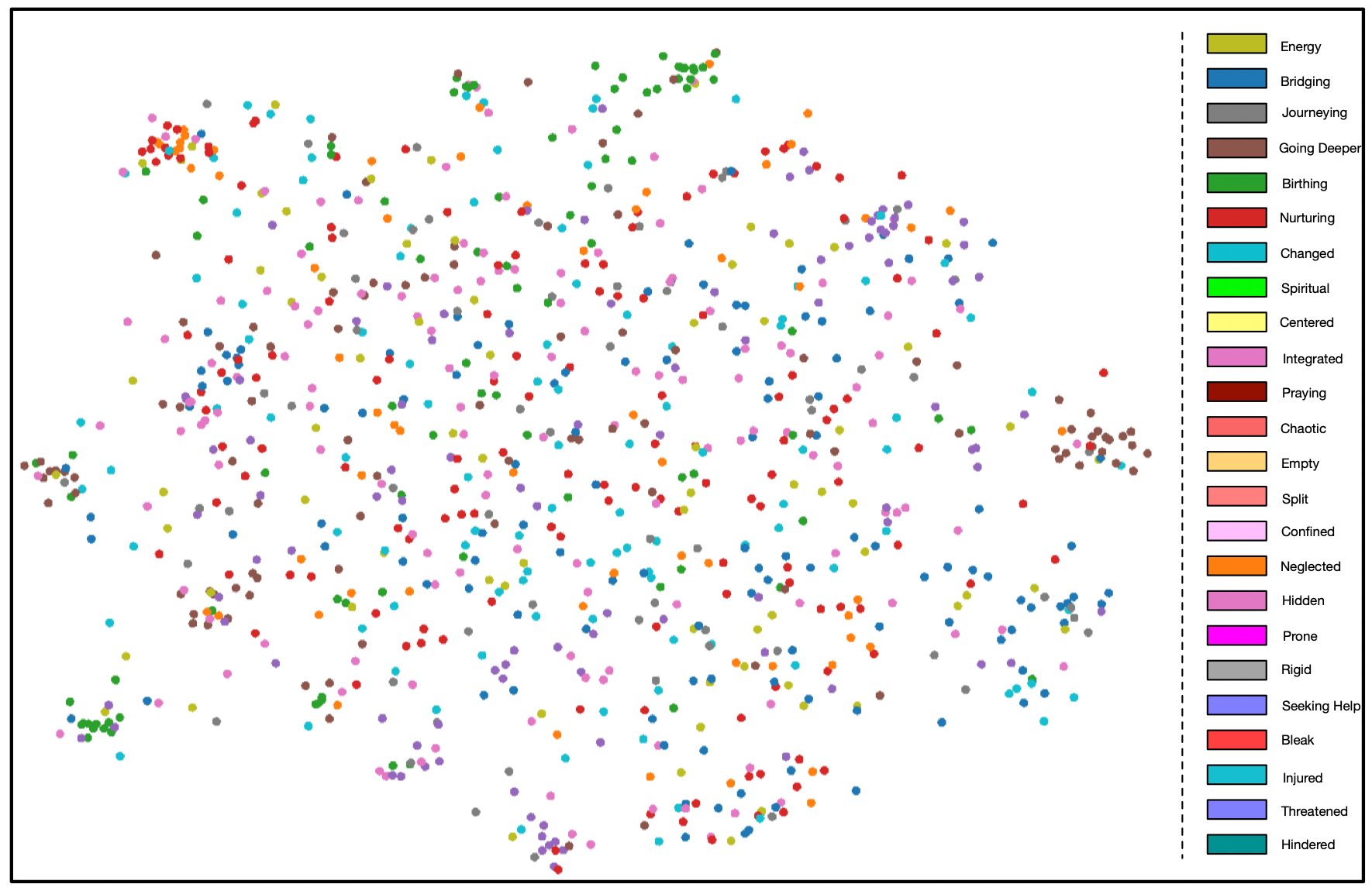} 
        \caption{\textbf{Relationship between psychological state categories and visual entity categories.} Different psychological state categories distribute evenly in space of visual entity categories. Indicating the psychological states relies weakly on local visual features, but may relies on further information.}
        \label{fig3}
    \end{figure}
    
    \section{4. Method}
    
    \subsection{4.1. Basic image captioning models}
    Building on SpyIn, applying image captioning approaches on VEIT is feasible. Borrow the idea from cutting-edge works, we deploy two image captioning approaches on SpyIn. These two approaches are based two popular backbone architectures respectively, which is representative for most of the approaches in image captioning. The first is a CNN-LSTM captioning approach, Show-Attend-Tell (SAT) \cite{SAT}, which consisting of an image encoder and a LSTM decoder with attention mechanism. The second is a state-of-the-art image captioning model, Meshed-Memory Transformers ($M^2$) \cite{M2}, which is a transformer-based image captioning approach relies on separately computed object-bounding-box detections. To avoid overfitting of transformer-based approach, we also present a downscaled model for comparison.
    
    \subsection{4.2. Visual-semantic model}
    For better serving VEIT, SpyIn dataset provides multimodal information using for optional input (See section 3). According to the dataset analysis above, interpretating psychological state may not perform well only relying on image features, and the results proves our suspicions (See section 5). As a professional task in psychology, psychological state interpretation may require profounder information input, such as scene graph related information and psychological knowledge information. Therefore, we perform an elementary but enlightening semantic combined model for VEIT.
    
    According to the result of baseline experiment (See section 5), we find a better performance from SAT architecture, which lead our semantic combined model designing to be mainly based on SAT. 
    We construct a dictionary including the scene graph related information (name, location coordinates, orientation of miniatures) and psychological knowledge related information (theme categories and attribute of miniatures). We use learnable embedding layer $\mathbf{W}_s$ to embed scene graph related information $\mathbf{D}_s$, and learnable embedding layer $\mathbf{W}_k$ to embed psychological knowledge information $\mathbf{D}_k$. Then we use a LSTM to encode these information into S:
    
    \begin{equation}\label{eq1}
        \mathbf{S} = {\rm LSTM}(\mathbf{W}_s\mathbf{D}_s;\mathbf{W}_k\mathbf{D}_k)
    \end{equation}
    
    On the other hand, image features V are extracted by a pretrained resnet-101. We concatenating $\mathbf{V}$ and $\mathbf{S}$ as a visual-semantic combined encoding, provide to decoder. The decoder is a LSTM based attention network for the sequence prediction of $\mathbf{C}$. Given a ground-truth interpretation $\mathbf{C}^*$ for the model, both the semantic encoder and the decoder is trained by minimizing the cross-entropy loss:
    
    \begin{equation}\label{eq2}
        \mathbf{L}_{XE} = - \log{P(\mathbf{C},\mathbf{C^*})}
    \end{equation}
    
    For ablation study, we set $\mathbf{W}_s$ to zero vector to achieve scene graph related information input only, and set $\mathbf{W}_k$ to zero vector to achieve psychological knowledge information input only.

    \begin{figure}[t]
        \centering
        \includegraphics[width=0.9\columnwidth]{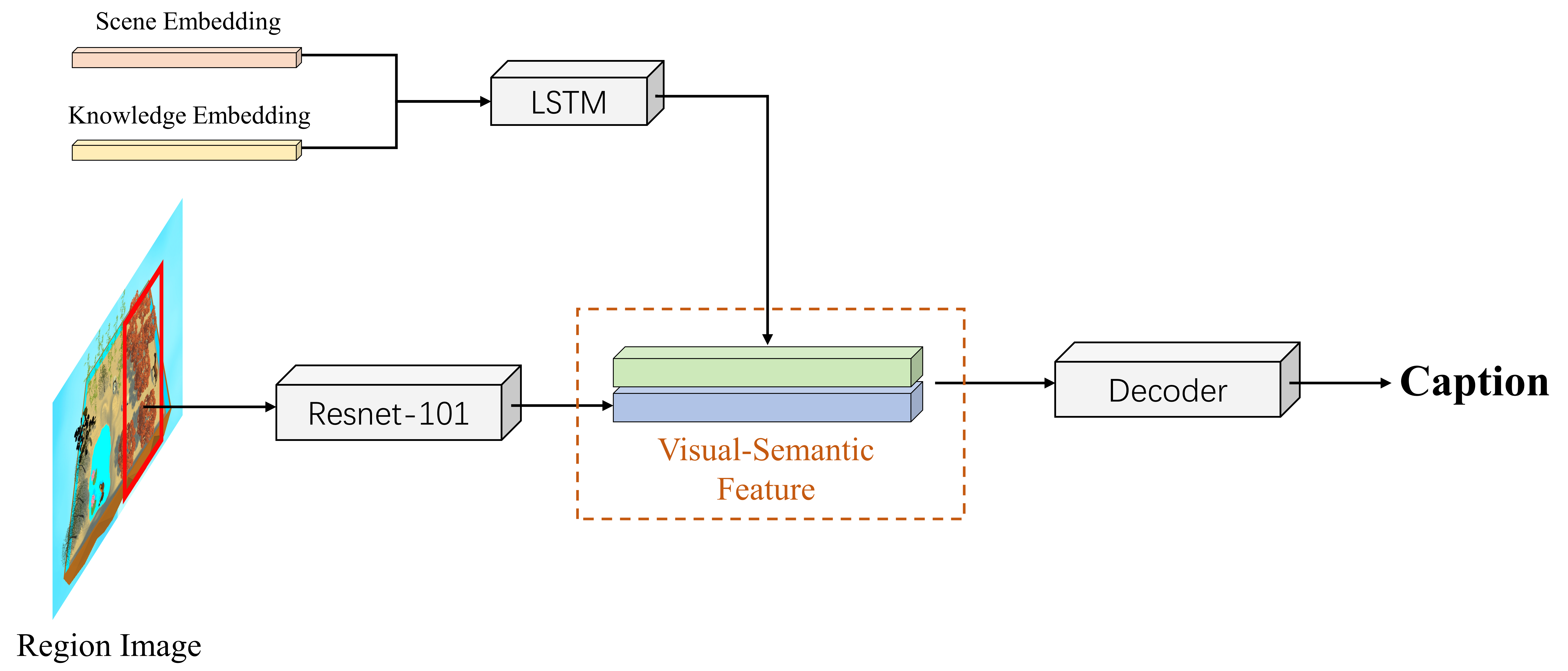} 
        \caption{\textbf{Architecture of visual-semantic combined model.} We combine the image features with scene graph information and psychological knowledge information embeddings.}
        \label{fig4}
    \end{figure}

    \subsection{4.3. Experimental Setup}
    We adopt an [85\%, 5\%, 10\%] train-validation-test data split on our dataset and do model-selection according to the performance on the validation split. 
    Both in the experiment of baseline model and visual-semantic model, we use the same backbone, pretrained resnet-101 for image feature extraction, and obtained 2048-dimensional feature vector for each image. We also use appropriate techniques like warmup and dropout to enhance the performance.
    In visual-semantic model, we use a 2 layer LSTM with 512 hidden nodes to encode semantic information. The encoder and decoder both adopt Adam as optimizer. The learning rate of encoder and decoder is $1 \times 10^{-4}$ and $4 \times 10^{-4}$ respectively.
    
    \subsection{4.4. Evaluation Metric}
    
    We adopt the interpretations annotated by expert psychotherapists as ground truth. To measure the generations are linguistically similar to ground-truth interpretations, we use various popular machine-based metrics: BLEU 1-4 \cite{BLEU}, ROUGE-L \cite{ROUGE}, METEOR \cite{MENTOR}, CIDEr-D \cite{CIDEr}.
    Notably, we bring BERTScore \cite{BERTScore}, a popular metric in natural language generation, into our evaluation. Corresponding to the motivation of serving psychoanalysis, evaluating the semantically similarity is more reasonable than word matching. BERTScore compute similarity using contextualized token embeddings in contrast to string matching, which fit better to VEIT.
    
    \section{5. Result}
    
    \begin{table}[h]
        \centering
        \resizebox{.90\columnwidth}{!}{
        \begin{tabular}{cccccc}
            \toprule
            Metric     &  $M^2$(Original/Downscaled)  &  SAT    &    SAT-S   \\
            \midrule
            BLEU-1     &  0.168 / 0.204 &  0.242  &   \textbf{0.295}    \\
            BLEU-2     &  0.091 / 0.107 &  0.127  &   \textbf{0.170}    \\
            BLEU-3     &  0.062 / 0.081 &  0.086  &   \textbf{0.122}    \\
            BLEU-4     &  0.046 / 0.062 &  0.064  &   \textbf{0.095}    \\
            ROUGE-L    &  0.205 / 0.230 &  0.288  &   \textbf{0.335}    \\
            METEOR     &  0.092 / 0.119 &  0.121  &   \textbf{0.150}   \\
            CIDEr-D    &  0.302 / 0.431 &  0.482  &   \textbf{0.874}    \\
            BERTScore  &  0.659 / 0.667 &  0.678  &   \textbf{0.711}    \\
            \bottomrule
        \end{tabular}
        }
        \caption{\textbf{Performance of different models on SpyIn.} $M^2$: Meshed-Memory Transformer (including a original one and a downscaled one), SAT: Show-Attend-Tell, SAT-S: Visual-semantic combined SAT.}
        \label{table3}
    \end{table}

    \begin{table*}[h]
        \centering
        \resizebox{.95\textwidth}{!}{
        \begin{tabular}{lcccccccc}
            \toprule
            \multicolumn{1}{c}{Features} & BLEU-1 & BLEU-2 & BLEU-3 & BLUE-4 & ROUGE-L & METEOR & CIDEr-D & BERTScore-F1 \\
            \midrule
            (i) Visual features & 0.242 & 0.127 & 0.086 & 0.064 & 0.288 & 0.121 & 0.482 & 0.678 \\
            + (ii) scene graph information only & 0.261 & 0.152 & 0.109 & 0.089 & 0.319 & 0.132 & 0.671 & 0.694 \\
            + (iii) psychological knowledge only & 0.239 & 0.136 & 0.102 & 0.078 & 0.295 & 0.129 & 0.632 & 0.688  \\
            + (iv) S\&K both & \textbf{0.295} & \textbf{0.170} & \textbf{0.122}  & \textbf{0.095}   & \textbf{0.335} & \textbf{0.150} & \textbf{0.874} & \textbf{0.711} \\
            \bottomrule
        \end{tabular}
        }
    \caption{\textbf{Alation study results.} We combine different extra information with visual features. S\&K both means that both scene graph information and psychological knowledge information are combined.}
    \label{table4}
    \end{table*}
    
    \begin{figure*}[h]
        \centering
        \includegraphics[width=0.8\textwidth]{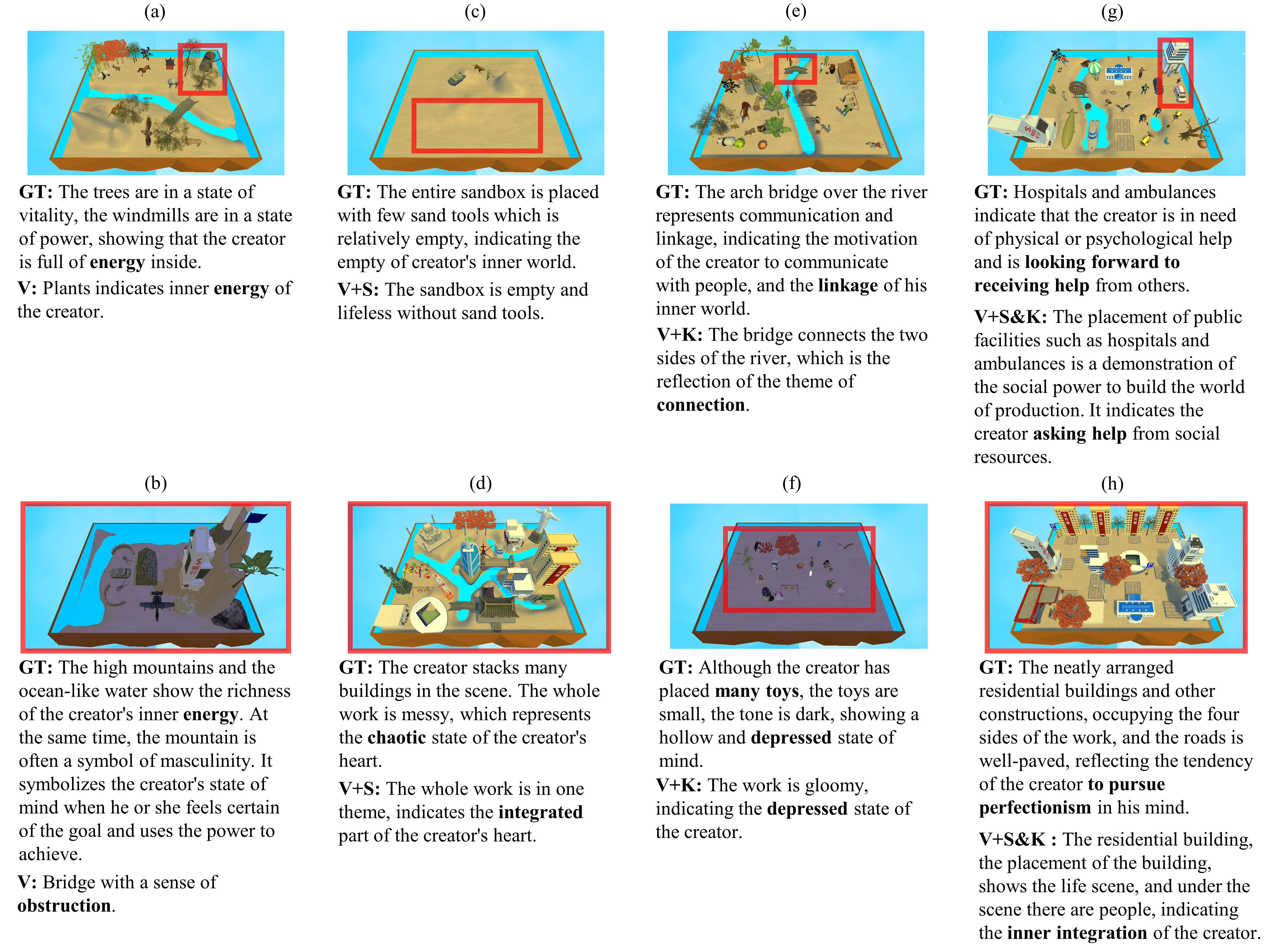} 
        \caption{\textbf{Example of interpretations generate by visual-semantic combined models.} The four columns are respectively generated by visual features only, viusal features combined scene graph information, visual features combined psychological knowledge information, visual features combined both two extra information.}
        \label{fig5}
    \end{figure*}
    
    \subsubsection{Result on image captioning method.}
    We find it is difficult to get a high numerical result in VEIT by simply migration image captioning models (See Table 3). Comparing to these architectures training and testing with COCO-captions (BLEU-1 with SOTA is 82.0 \cite{M2}), the result with SpyIn is noticeable worse. However, the result is expected. According to section 3, the annotations of SpyIn is specifically linguistic richer and more abstract. Moreover, according to the statistic in section 3.3, VEIT is not only a simple visual-pattern-relied task, but need further information or external knowledge to assist. The result of the visual-semantic combined model validated our assumptions. The SAT-S model attain improvement in every indicator and especially a huge improvement of 81.3\% in CIDEr-D. Semantic combined model apparently improves the result shows a great independence of VEIT on external information.
    
    The transformer-architecture-based approach M2 doesn't get a better score than SAT. We suppose the reason is that the data volume of SpyIn is smaller than COCO-caption, which may lead to overfitting. Therefore, we downscaled the hidden sizes of origin M2 model, and operate another comparison experiment which shows a better result to origin model. But the score is still lower than SAT. The result we get is very similar to previous work \cite{itsokay}, which indicating SAT may obtain a better result than $M^2$ on non-photographic dataset.

    \subsubsection{Result on semantic combined method.}
    We conduct the ablation study to find which kind of information contributes more to VEIT. Both scene-graph related information and psychological knowledge related information improves the scores in most of the metrics. The input of scene-graph related information obtains an improvement of 39.2\% in CIDEr-D and 2.4\% in BERTScore. It indicates the VEIT relies not only on visual features but also on macroscopic scene information, for example the location relationship between visual entities. The psychological knowledge related information doesn't get a improvement on BLEU-1, but attains an improvement of 31.1\% on CIDEr-D and 1.5\% on BERTScore. It indicates the psychological knowledge related information may not directly improve the word-for-word result, but do help to the meaning globally. The result of ablation study shows the strong dependency of both scene and knowledge information for VEIT. (The interpretating result of different models is shown in Figure 5.)

    \section{6. Conclusion}
    Meeting the demands of psychoanalysis, it is a challenging work for AI to interpret creators' psychological states through visual creations. In this work, we take the first step to define VEIT and present SpyIn, a psychological theory supported and professional annotated dataset, to serve VEIT. We adopt several captioning approaches to perform VEIT on SpyIn, and the results show that VEIT is a challenging task reling not only on visual features but also on further knowledge. The results of visual-semantic combined approach show the promise for AI to analyze and articulate the inner world of humanity through our visual works.
    
    \appendix
    \bibliography{aaai23}

\end{document}